\RequirePackage{fix-cm}
\documentclass[smallextended]{svjour3}       
\smartqed  
\usepackage{graphicx,amssymb,amsmath,amsfonts,subfigure, algorithm, algorithmic, multirow, amsmath, xcolor, graphics, graphicx, epsfig, multirow,bm,epsfig,rotating}
\begin{document}

\title{Multi-view Point Cloud Registration with Adaptive Convergence Threshold and its Application in 3D Model Retrieval
}


\author{Yaochen Li $^{1}$        \and
        Ying Liu   $^{1}$       \and
        Rui Sun $^{1}$  \and
        Rui Guo $^{1}$ \and
        Li Zhu  $^{1}$  \and
        Yong Qi $^{2}$ 
}


\institute{1 School of Software Engineering, Xi'an Jiaotong University, Shaanxi, China\\
              \email{yaochenli@mail.xjtu.edu.cn}            \\
           2 Department of Computer Science, Xi'an Jiaotong University, Shaanxi, China
}

\date{Received: date / Accepted: date}

\maketitle

\begin{abstract}
Multi-view point cloud registration is a hot topic in the communities of artificial intelligence (AI) and multimedia technology. 
In this paper, we propose a framework to reconstruct the 3D models by the multi-view point cloud registration algorithm with adaptive convergence threshold, and apply it to 3D model retrieval subsequently. 
The iterative closest point (ICP) algorithm is implemented with adaptive convergence threshold, and further  combines with motion average algorithm for the registration of multi-view point cloud data. 
After the registration process, the applications are designed for 3D model retrieval. 
The geometric saliency map is computed based on the vertex curvatures. 
The test facial triangles are selected to compare with the standard facial triangle. 
The face and non-face models are then discriminated. 
The experiments and comparisons prove the effectiveness of the proposed framework.
\end{abstract}

\section{Introduction}
\label{intro}
Artificial intelligence (AI) and multimedia are important technologies to support people's daily  life and economic activities \cite{Lu_HM}. 
These technologies can be applied to the applications of cross-modal retrieval \cite{Xu_X}, motor anomaly detection \cite{Li_YJ}, etc. 
Multi-view point cloud registration is a hot aspect in the AI and multimedia community.
With the rapid development of laser scanning technology, it is possible to obtain high precision data of the object surface \cite{Chane_CS}. 
Due to the limited scanning range, the laser scans usually cannot obtain complete 3D object information. 
Therefore, it is important to effectively integrate the point cloud data collected from different views to generate a complete model.
In the traditional ICP algorithms, the convergence threshold is mostly manually set, which demands too much time cost and easily lead to the inaccurate registration results.
The adaptive convergence threshold proposed in this paper will reduce the manual participation, thus automate the registration algorithm. 
The reconstructed 3D models can be applied to the applications of 3D model retrieval, 3D textured model encryption \cite{Xin_Jin}, 3D point cloud encryption \cite{Xin_Jin2}, etc. \par   

  Pair-wise registration is the basis of multi-view registration. 
Iterative closest point (ICP) algorithm is a classic algorithm for the pair-wise registration, which aims to calculate the optimal space transformation between a pair of point cloud data. 
However, this algorithm has the drawbacks of local convergence, and cannot solve the problem of partial overlap  between point cloud pairs. 
Chetverikov et al. \cite{Chetv_D} improve the ICP algorithm by utilizing the overlap rate for effective selection, namely trimmed iterative closet point (TrICP) algorithm.
Lomonosov et al. \cite{Lomonosov_E} implement the genetic algorithm for global search to get the optimal registration of initial value. 
Sandhu et al. \cite{Sandhu_R} apply particle filter to make the registration of the initial value more effective, and improve the ICP algorithm with global convergence properties. 
However, these methods can only deal with the registration between point cloud pairs.\par

Multi-view registration of point cloud data can be implemented on the basis of pair-wise registration, which is  more challenging for more parameters need to be considered \cite{Zhu_JH,Guo_R}. 
The multi-view registration problem can be solved by quadratic programming of Lie algebra parameters \cite{Shi_SW}, where each node and edge in the graph denotes a point cloud and the related pairwise registration.
Chen et al. \cite{Chen_Y} propose the basic concept of point cloud registration, which matches the adjacent point clouds by the ICP algorithm, and transform all the point cloud data into a single global coordinate system. However, this method can be easily influenced by the local convergence property of the ICP algorithm. 
Bergevin et al. \cite{Bergevin_R} propose an improved ICP algorithm to deal with the rotations and transformations among all the point cloud data. 
However, this method is computationally expensive, especially in the big data case.
Guo et al. \cite{Guo_YL} and Fantoni et al. \cite{Fantoni} propose approaches for the registration of point cloud data by extracting the vertex features.
However, the registration failure occurs in the case that insufficient features are obtained.\par

As an application based on the multi-view registration of point cloud data, 3D face retrieval is  a hot issue in the community of 3D model computation. 
In the studies of \cite{Akagunduz_E}, a generic method for 3D face detection and modeling is proposed. 
The multi-scale analysis is implemented by computing Gaussian mean curvatures. 
Creusot et al. \cite{Creusot_C} present an automatic mechanism to detect key points on 3D faces, which constitutes a local shape dictionary. 
In the studies of Rabiu et al. \cite{Rabiu_H}, a face segmentation method is presented with adaptive radius. 
The intrinsic properties of face are derived from Gaussian mean curvature for segmentation. 
The similar methods are described in the studies of \cite{Lei_J,Boukamcha_H}.\par

In this paper, a multi-view point cloud registration method with adaptive convergence threshold is presented. 
The proposed method improves the classic ICP algorithm with adaptive convergence threshold and combines the motion average algorithm. 
The reconstructed 3D model is applied to the application of 3D model retrieval. 
The 3D face detection is implemented by clustering the geometric saliency into facial triangles.
The face and non-face models are disciminated by the dissimilarity of test and standard facial triangles.
The main contribtions of this paper are summarized as follows: 
\begin{itemize}
\item The adaptive convergence threshold is applied to the ICP algorithm, which makes the registration of multi-view point cloud data more effectively; 
\item The motion average algorithm is utilized for multi-view point cloud registration, thus the error accumulation problem can be solved.
\item A new 3D face detection method is proposed based on the matching of test and standard facial triangles, using the geometric saliency of surface.
\end{itemize}

The remainder of the paper is orginized as follows: In Section 2, the multi-view point cloud registration algorithm with adaptive convergence threshold is introduced. Section 3 describes the application of 3D model retrieval based on the multi-view point cloud registration results. The experiments and analysis are conducted in Section 4, followed by the conclusion and future works described in Section 5.\par

\section{Multi-view Point Cloud Registration with Adaptive Convergence Threshold}

In this section, we propose the algorithm of multi-view point cloud registration with adaptive convergence threshold. 
Firstly, the concept of adaptive convergence threshold is introduced. 
The ICP algorithm with adaptive convergence threshold and the motion average algorithm are described subsequently. Finally, the proposed multi-view registration algorithm is presented.

\subsection{Adaptive Convergence Threshold}

For the array point data, the horizontal distance between the 3D points can be represented by the horizontal resolution. 
The traditional laser point cloud lacks the parameter of horizontal resolution, and the horizontal distance between the 3D points can be estimated by the average distance of 3D points.
The public data sets are mostly based on the laser point cloud rather than the array point cloud, so we mainly use the laser point cloud in this paper. 
The array point cloud will be studied in the future.
One of the stopping conditions for the pair-wise registration of point cloud data is the point set distance. 
The registration of two point cloud sets is implemented if the point set distance is lower than a certain threshold. 
The distance of point sets will reach minimum value if they get close. 
As a result, we predict the ideal distance of point sets for registration, which is utilized as the distance threshold.\par

According to the studies of Wang et al. \cite{Wang_Y}, each ideally matched corresponding point in the data set that is located inside a circle with radius of $\sqrt{2}L_r/2$  around the modified point, where $L_r$ is the horizontal resolution. 
Moreover, the distance between the point pair will be influenced by  the overlap rate. 
The average distance between the point pair decreases with the increase of the overlap rate.
Thus we utilize the weighted overlap rate to make the distance more reasonable.
At the same time, the distance error is another a factor that influences the registration result.
The weight of the overlap rate can also be considered, and the registration convergence threshold is expressed by:
\begin{equation}
{e_{thr}} = {[(1 - \frac{{{N_p}}}{{{N_t}}})\frac{{\sqrt 2 }}{2}{L_r}]^2} + {(\frac{{{N_t}}}{{{N_p}}}{R_e})^2}
\end{equation}
where $N_p$ denotes the number of coincidence points, $N_t$  is the total number of point cloud data. $R_e$ indicates the range accuracy of the system.\par
In this paper, the laser point cloud is utilized instead of the array point cloud. As a result, the above equation can be modified to some extent.\par
\begin{itemize}
  \item The parameters to denote the horizontal resolution in the laser scanning system are lacked. 
The distance between the point cloud data can be considered to follow the uniform distribution, since the uniform scanning is adopted. 
As a result, the uniform distance between the point cloud data is computed as the parameter of horizontal resolution.
  \item The point cloud data are selected before registration.
Only the point cloud data with the overlap rate more than 50\% will be considered.
As a result, the overlap rate has little influence on the system ranging.
Thus we remove the overlap rate for the weighted system ranging.
\end{itemize}

The modified convergence threshold is defined as follows:
\begin{equation}
{e_{thr}} = {[(1 - \frac{{{N_p}}}{{{N_t}}})\frac{{\sqrt 2 }}{2}{L_r}]^2} + {({R_e})^2}
\end{equation}

\subsection{ICP Algorithm based on Adaptive Convergence Threshold}

\begin{figure}[!htbp]
  \includegraphics[width=9cm]{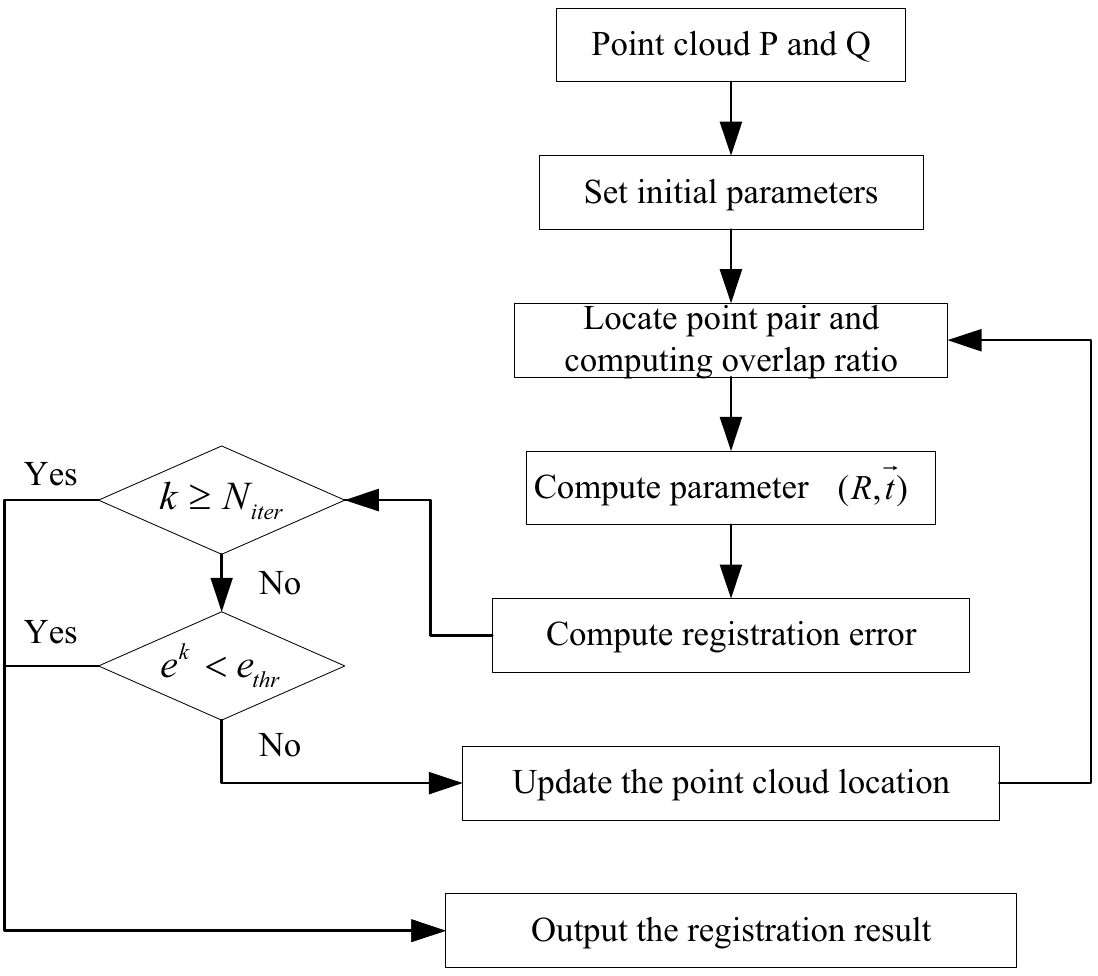}
\caption{Flow diagram of ICP algorithm with adaptive convergence threshold.}
\label{fig:Flow_diagram}       
\end{figure}

Given the point cloud data $P$ and $Q$, we denote $R \in {R^{3 \times 3}}$  as the rotation parameter, and $\vec t \in {R^3}$ as the translation parameter. 
The flow diagram of the ICP method with adaptive convergence threshold is shown in Fig. \ref{fig:Flow_diagram}, which mainly includes the following steps:
\begin{itemize}
  \item Localization of the point cloud pairs. Firstly, we perform the transformation for the point cloud $P$ using the initial parameters of rotation and transformation acquired from the rough registration. 
Secondly, the nearest point in $Q$ is searched for each point in $P$.
The overlap rate is computed simultaneously.

  \item Computation of the parameter $(R,\vec t)$. 
The total distance of all the corresponding points is computed. 
Furthermore, the optimal rigid transformation $(R,\vec t)$ is specified, which can minimize the sum of distances.

  \item Transformation of the point cloud $P$. 
The point cloud $P$ is transformed based on $(R,\vec t)$  obtained from the second step.

  \item Registration of the point cloud data. The iteration computation terminates at two conditions: 
(1) The average distance between the newly transformed point cloud and the model point cloud is less than the adaptive threshold $e_{thr}$;
(2) The maximum iteration number $N_{iter}$  is reached. Otherwise, the newly transformed point cloud is utilized as a new point cloud $P$ to continue the iteration.
\end{itemize}
\par The ICP algorithm with adaptive convergence threshold makes the registration of point cloud data more accurately.
After filtering the unreasonable corresponding points, the transformation relationship between the point cloud data.
Furthermore, more accurate registration results and less matching errors are reached.
However, this algorithm only deals with the registration of the pairwise point cloud data.
Thus the motion average algorithm is implemented for the multi-view point cloud registration.

\subsection{Motion Average Algorithm}
The 3D model constructed directly from multi-view point cloud registration will lead to error accumulation.
The motion average algorithm aims to solve the error accumulation problem, which computes the error correction terms by comparing the precise relative motion and the approximate relation motion.
The error correction terms are converted from the Lie group to the Lie algebra.
The column vectors are then extracted from the Lie algebra space.
Subsequently, the error correction information is assigned to the global motion of each point cloud to alleviate the error accumulation.
The main steps of the motion average algorithm are as follows:\par

\begin{itemize}
  \item (1) Compute the correction value between the relative motion  $M_{ij}$ and the initial relative motion  $M_{ij}^0$:
\begin{equation}
\Delta {M_{ij}} = M_i^0{M_{ij}}{(M_j^0)^{ - 1}}
\end{equation}
                                       
  \item (2) The correction value is transformed from the Lie group $M$  to the corresponding Lie algebra $m$ \cite{Shi_SW}:
\begin{equation}
\Delta {m_{ij}} = \log (\Delta {M_{ij}})
\end{equation}                                   
  \item (3) The column vector is established by the Lie algebra matrix:
\begin{equation}
\Delta {v_{ij}} = vec(\Delta {m_{ij}})
\end{equation}                                                   
  The matrix with the correction information for global motion is obtained by computing the column vector  $\Delta {V_{ij}}$, which contains all the error correction terms, as well as the matrix $D$ for the point cloud relations.
\begin{equation}
\Delta {V_{ij}} = [\Delta {v_{ij1}},\Delta {v_{ij2}},...,\Delta {v_{ijn}}]
\end{equation}
\begin{equation}
\Delta \Im  = {D^ \dagger }\Delta {V_{ij}}
\end{equation}
  \item (4) Extract the column vectors of each point cloud in matrix $\Delta \Im$, and restore the Lie algebra form:

\begin{equation}
\Delta \Im  = {[\Delta {v_1},\Delta {v_2},...,\Delta {v_k},...,\Delta {v_N}]^T}
\end{equation}
\begin{equation}
\Delta {m_k} = cev(\Delta {v_k})
\end{equation}
                                   
  \item (5) The Lie algebra is converted into the Lie group for each point cloud, which is applied to rectify the global motion of each frame:
\begin{equation}
\forall k \in [2,N],{\kern 1pt} {\kern 1pt} {\kern 1pt} {M_k} = {e^{\Delta {m_k}}}{M_k}
\end{equation}
                                    
  \item (6) For the newly computed global motion of each point cloud, we perform $M_{global}^0 = {M_{global}}$. Repeat steps (1)-(6) until all the column vectors with error correction term delta $\Delta \Im$  is relatively small, i.e. $
\left\| {\Delta \Im } \right\| < \varepsilon $.

\end{itemize}

The motion average algorithm eliminates the accumulated errors, which can get more accurate rotation translation matrix.
As a result, more accurate results for multi-view point cloud registration will be achieved.

\subsection{Multi-View Registration Algorithm}
The input of the algorithm mainly include the point cloud data and the initial global motion of $N$ frames, while the output are the final precise global motion. The main steps are as follows: (1) The overlap rate is computed between pairwise point cloud data; (2) If the computed overlap rate in (1) is greater than $
{\xi _0}$, the pair-wise registration is proceeded. 
The ICP algorithm with convergence threshold is implemented to get the relative motion; 
(3) The more accurate global motion is computed by motion average, and the corresponding error is computed at the same time; 
(4) If the computed error is less than a certain threshold, the iteration stops; Otherwise, the current global motion is utilized as a new initial value.\par
  The process of multi-view registration is summarized in Algorithm \ref{Alg:Point_cloud}.
The function $comput\_overlap(P_i,P_j)$  represents the overlap percentage between the  point cloud $p_i$ and $p_j$.
Function $testICP(P_i, P_j)$  denotes the relative motion between point cloud  $P_i$ and $P_j$  using the ICP algorithm based on adaptive convergence threshold. 
The function $Motion\_Averaging(M_{ij1},M_{ij2},...,M_{ijn})$  is the relative motion of a sequence of non-adjacent point clouds as input. 
The motion average algorithm is implemented to compute the accurate global motion between each point cloud. 
$R_i$ and $R_i^0$ denote the rotation matrices of adjacent iterations, while the function $e^k=cal\_error(R_i,R_i^0)$ represents the matching error between the adjacent iterations.

\begin{algorithm}[!htbp]
\caption{Multi-View Point Cloud Registration}
\begin{algorithmic}[1]
\REQUIRE $\begin{cases}
\text{$\{P_1,P_2,...,P_N\}$}\\
\text{$M_{global}^0 = \{ {M_i},{M_j},...,{M_N}\} $}\\
\end{cases}$
\WHILE {$
iter < K{\kern 1pt} {\kern 1pt} {\kern 1pt} {\kern 1pt} \& {\kern 1pt} {\kern 1pt} {\kern 1pt} {\kern 1pt} {e^k} > {e_{thr}}$}
\FOR {$i=1:N$}
\FOR {$
j = 1:N{\kern 1pt} {\kern 1pt} {\kern 1pt} \& {\kern 1pt} {\kern 1pt} {\kern 1pt} j \ne i$}
\STATE    $
{\xi _{ij}} = comput\_overlap({P_i},{P_j})$\\
\IF {$
{\xi _{ij}} > {\xi _0}$}
\STATE $
{M_{ij}} = testICP({P_i},{P_j})$

\ENDIF
\ENDFOR
\ENDFOR

\STATE $
{M_{global}} = Motion\_Averaging({M_{ij1}},{M_{ij2}},...,{M_{ijn}})$;

\STATE $
{e^k} = cal\_error({R_i},R_i^0)$;

\STATE $iter=iter+1$;

\STATE $
M_{global}^0 = {M_{global}}$;

\ENDWHILE

\ENSURE $
M_{global}^{}:\{ I,{M_2},...,{M_N}\} $
\end{algorithmic}
\label{Alg:Point_cloud}
\end{algorithm}

\section{The Application of 3D Model Retrieval}

The multi-view registration of point cloud data can be applied to many useful applications, such as 3D model retrieval, 3D model reconstruction, etc. 
The 3D models are more intact based on the registration results, and thus has more geometric features for model retrieval. 
In this section, we use 3D face model as an example to illustrate the 3D model retrieval process. The model retrieval process mainly includes two steps: (1) Geometric saliency computation, and (2) facial triangle match.

\subsection{Geometric Saliency Computation}

Given the mean curvatures of discrete vertices, the geometric saliency of a 3D face model is represented by  Laplace-Beltrami operator to aggregate the 3D model vertices \cite{Lee_CH}.\par

  We denote  $N(v,\sigma ) = \{ x|\left\| {x - v} \right\| < \sigma \} $ as the collection that the vertices are within the distance  $\sigma$ to vertex $v$. 
The weighted Gaussian curvature of the vertex $v$ can be computed by:
\begin{equation}
G(l(v),\sigma ) = \frac{{\sum\limits_{x \in N(v,\sigma )} {l(x)\exp [ - {{\left\| {x - v} \right\|}^2}/(2{\sigma ^2})]} }}{{\sum\limits_{x \in N(v,\sigma )} {\exp [ - {{\left\| {x - v} \right\|}^2}/(2{\sigma ^2})]} }}
\end{equation}        	          
where $l(x)$  is the average curvature of vertex $x$.\par
  In order to compute the geometric saliency, the vertex saliency at different scales are computed by:

\begin{equation}
{\varphi _i}(v) = \left| {G(l(v),{\sigma _i}) - G(l(v),2{\sigma _i})} \right|
\end{equation}                              
where $\sigma _i$  is the standard deviation of the Gaussian filter at scale $i$, and the scale values in the experiment are $\{ 2\xi ,3\xi ,4\xi ,5\xi ,6\xi \}$ . $\xi$  is set to be 0.3\% of the diagonal length of the model surround box.
  It is necessary to cluster the geometric saliency regions after obtaining the geometric saliency map. Firstly, we set the saliency threshold as $th_{saliency}$  to get the high saliency regions. Secondly, the distance threshold $th_{dist}$  is specified. The saliency regions where the distance is less than the distance threshold are considered as the same cluster.

\subsection{Triangle Match and Error Computation}
\begin{figure}[!htbp]
  \includegraphics[width=9cm]{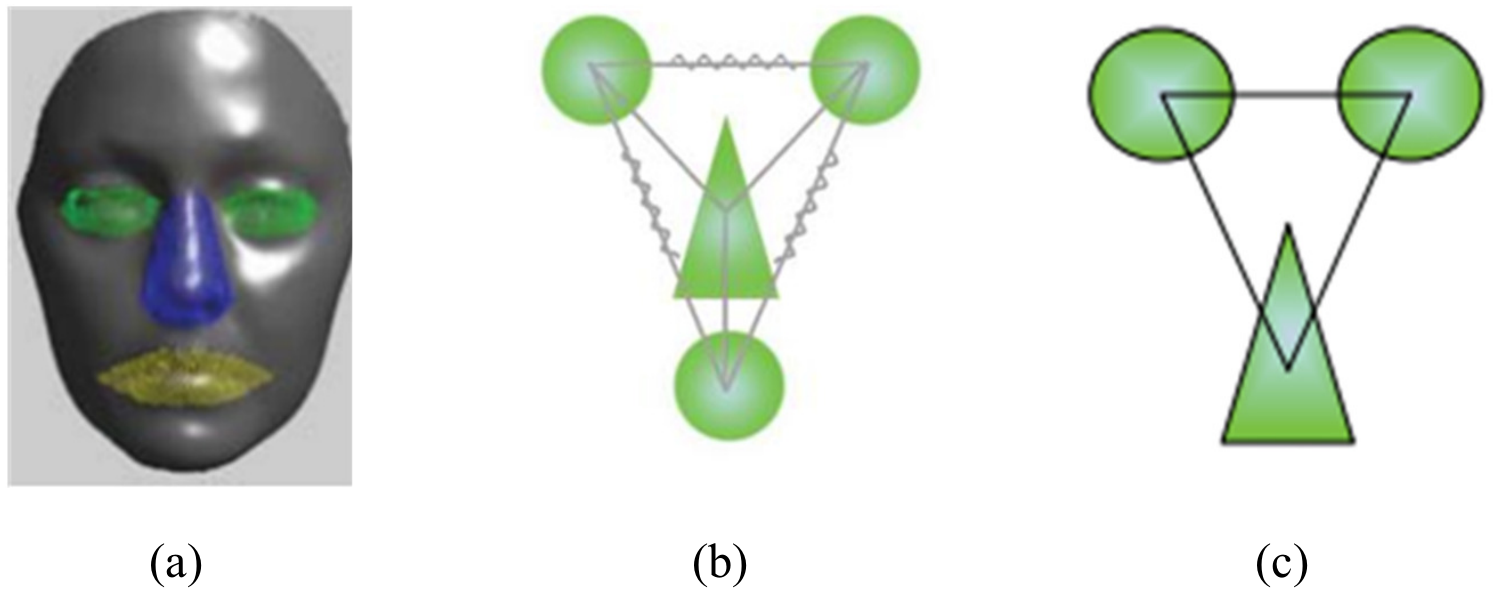}
\caption{Illustration of the facial triangle. (a) Vertex sets of the facial salient regions. (b)Topology of the 3D face model. (c) Facial triangle model.}
\label{fig:facial_triangle}       
\end{figure}

Given three salient regions, a triangle will be formed by saliency clustering.
Assuming $N$ salient regions exist in the 3D facial mesh, then we will select  $C_N^3$ facial triangles according to permutation and combination. 
The eye and nose regions of 3D face have more regional features, so the standard facial graph model is composed of the eye and nose regions.
The illustration of the facial triangle is shown in Fig. \ref{fig:facial_triangle}.\par

According to the features of the 3D face model, the registration between the standard and test facial triangles is performed, where the standard facial triangle $FT_S$ is annotated based on the standard 3D face model $D_S$.
The following steps are implemented for the registration: 
(1) The test triangle is shifted to make its center coincide with that of the standard facial triangle; 
(2) The test triangle is rotated to make its normal vector consistent with that of the standard facial triangle; (3) The test triangle is scaled to make its total vertex distance to the standard triangle reach the minimum. 
If the matching error between the test and the standard triangles is less than the threshold $Thres_{FT}$, the 3D face model is confirmed.
The 3D face model retrieval process is summarized in Algorithm \ref{Alg:3D_face}.

\begin{algorithm}[!htbp]
\caption{3D face model retrieval}
\begin{algorithmic}[1]
\REQUIRE $\begin{cases}
\text{Test 3D model dataset $\{D_1,D_2,...,D_k\}$}\\
\text{Standard 3D face model $D_S$ and its standard facial triangle $FT_S$}\\
\end{cases}$

\FOR {$i=1:k$}

\STATE  Compute vertex curvatures for the test 3D face model $D_i$;

\STATE  Select $C_N^3$ facial triangles based on the geometric salienct regions of $D_i$;

\FOR {$j=1:C_N^3$}

\STATE Perform the registration between the $j_{th}$ facial triangle of $D_i$ and $FT_S$, and compute the matching error $E_{FT}$;

\IF {$E_{FT}<Thres_{FT}$}
\STATE Flag=true;
\STATE Continue;
\ENDIF

\ENDFOR
\ENDFOR

\ENSURE 3D face models in the test 3D model dataset.
\end{algorithmic}
\label{Alg:3D_face}
\end{algorithm}

\section{Experiments and Analysis}
\begin{figure}[!htbp]
  \includegraphics[width=10cm]{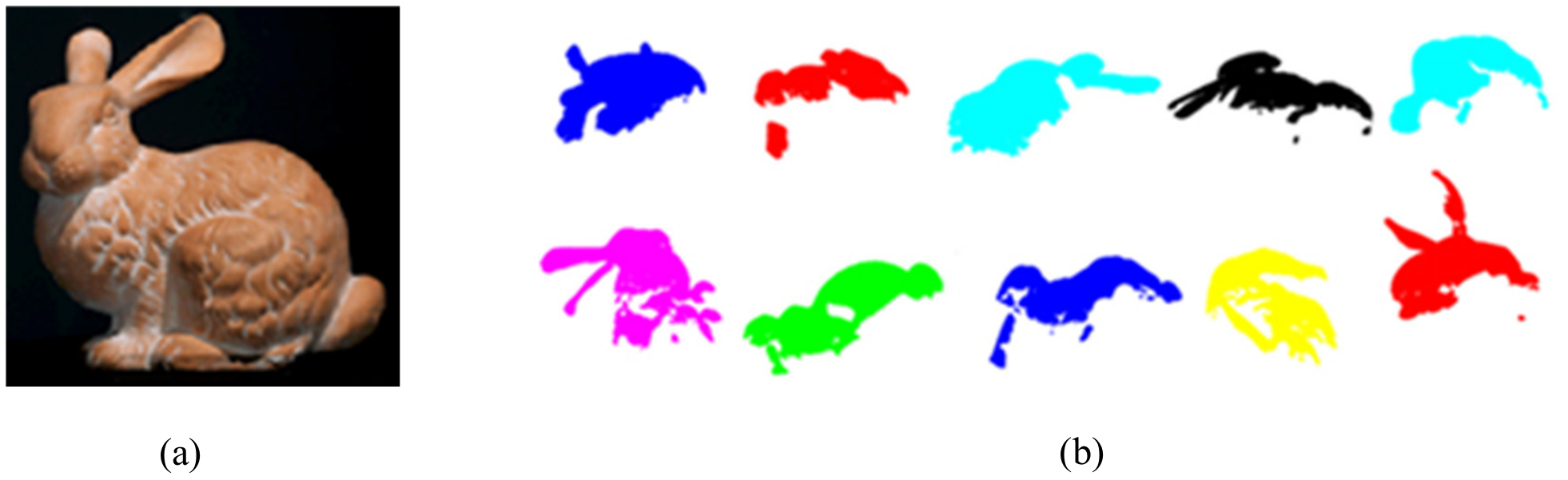}
\caption{(a)Bunny model. (b) The model data plotted from different views.}
\label{fig:Bunny_model}       
\end{figure}

\begin{figure}[!htbp]
  \includegraphics[width=10cm]{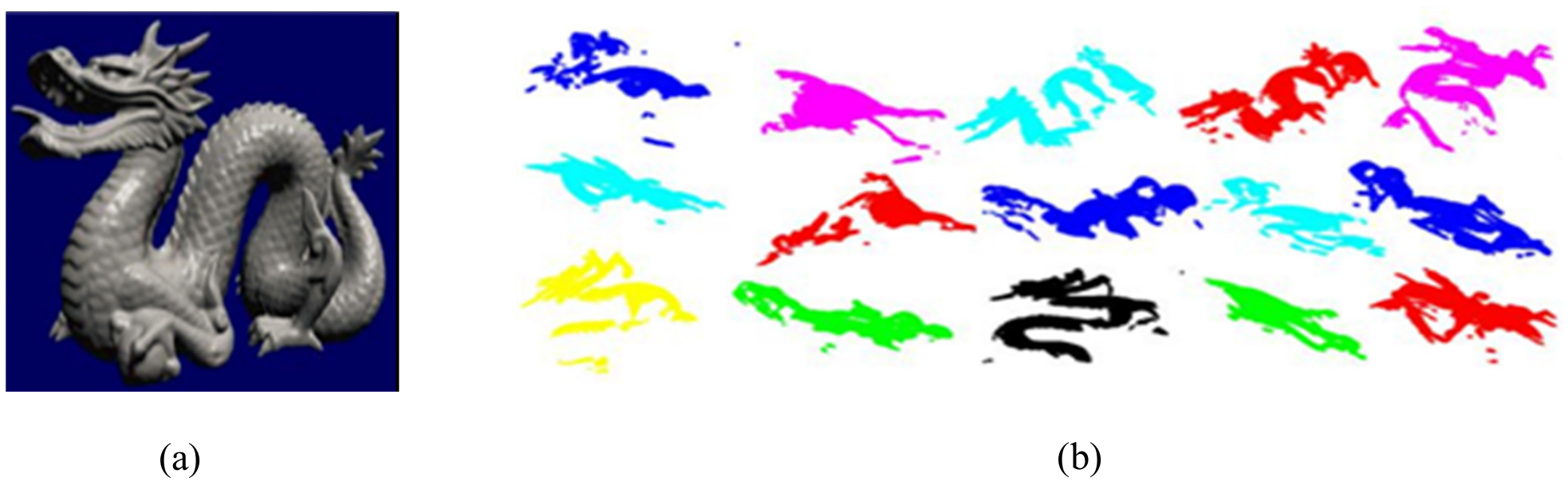}
\caption{(a)Dragon model. (b) The model data plotted from different views.}
\label{fig:Dragon_model}       
\end{figure}

\begin{figure}[!htbp]
  \includegraphics[width=10cm]{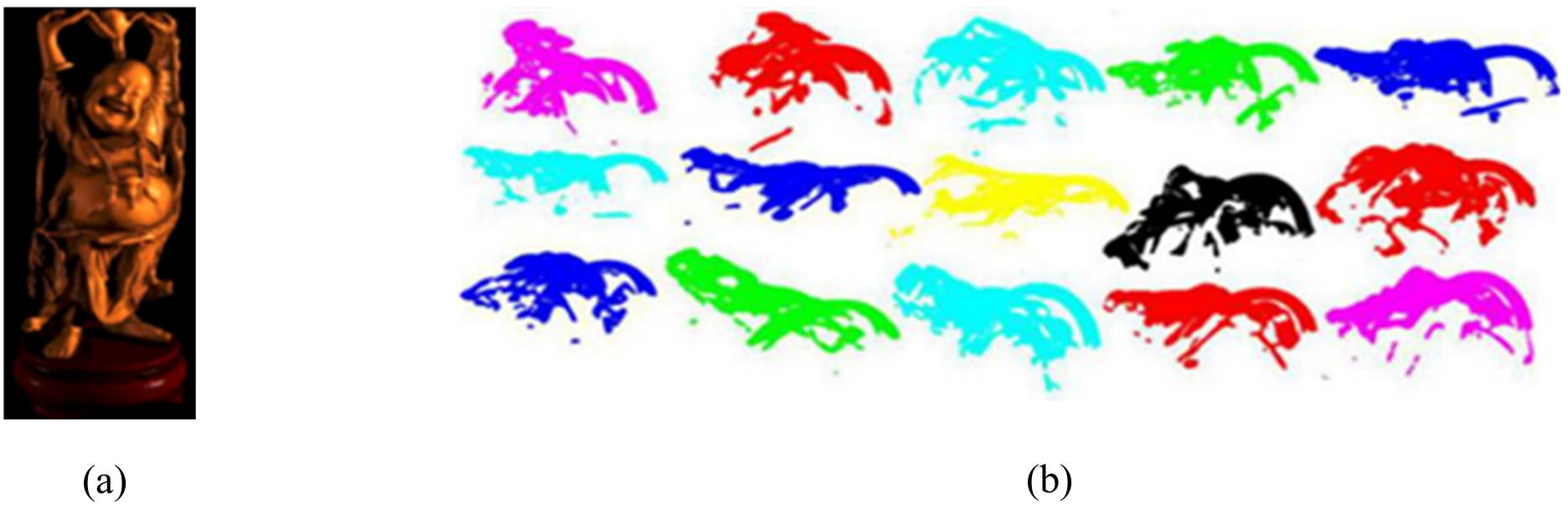}
\caption{(a)Happy model. (b) The model data plotted from different views.}
\label{fig:Happy_model}       
\end{figure}

In this section, the experiments are designed to evaluate the proposed algorithms. 
The experimental data for multi-view point cloud registration are collected from the Stanford 3D Model Database \cite{Stanford}, as shown in Fig. \ref{fig:Bunny_model}, Fig. \ref{fig:Dragon_model} and Fig. \ref{fig:Happy_model}. \par

\begin{figure}[!htbp]
  \includegraphics[width=8cm]{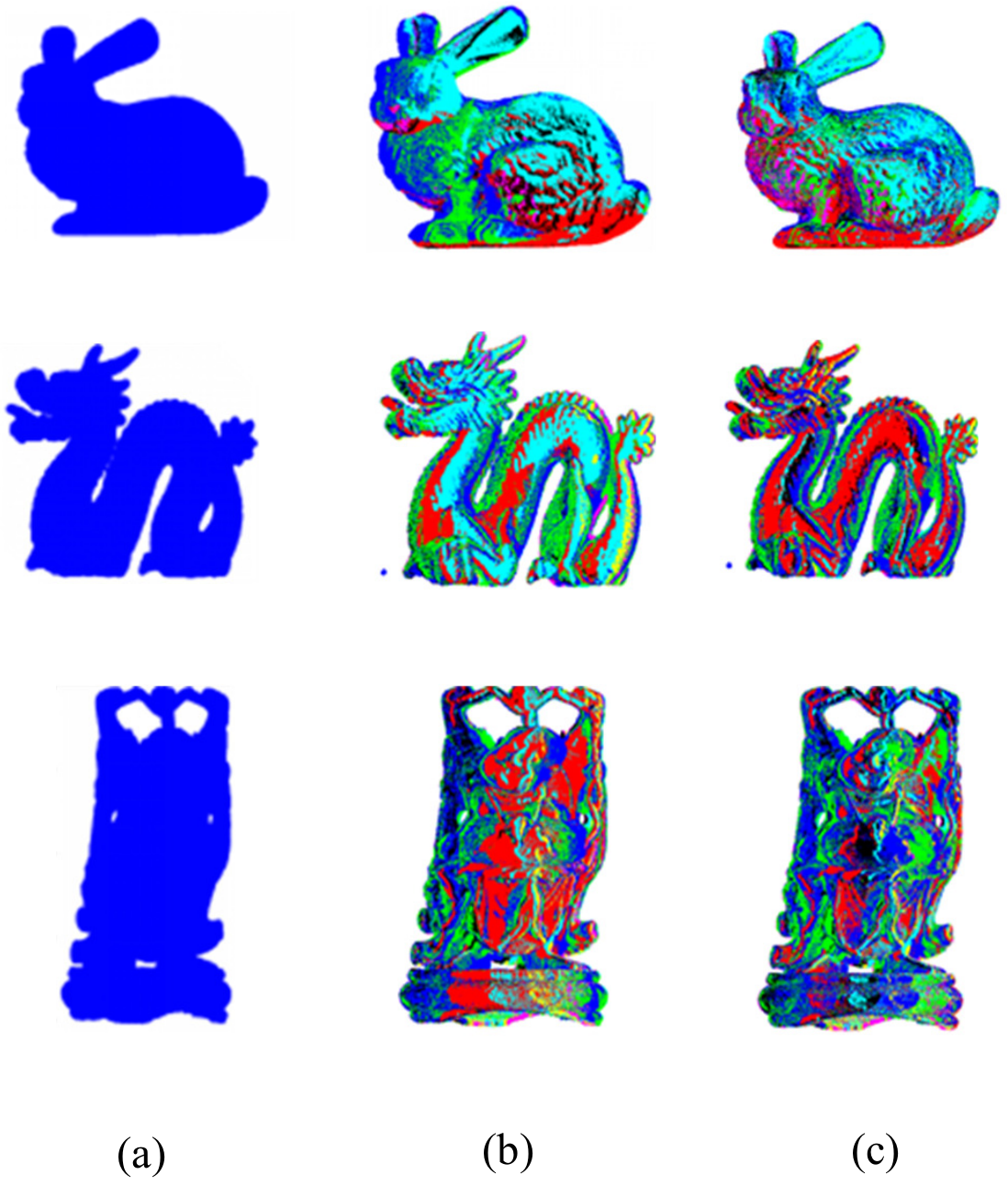}
\caption{Comparison of models in mottled degree. (a)3D model. (b)Initial model after rough registration. (c) Final model after multi-view registration.}
\label{fig:Mottle_degree}       
\end{figure}

Firstly, we compare the sectional views for the rough and precise registration results of multi-view point clouds, as shown in Fig. \ref{fig:Mottle_degree}. The comparison results of mottled degree prove that the registration of point clouds is most accurate after precise registration. 
The results of the low-rank and sparse matrix decomposition (LRS-Llalm), motion average based on ICP(MA-ICP) and the proposed method(MA-ATICP) are similar in the metric of mottled degree, so we apply original data and the cross section of them to illustrate the experimental results.\par

To quantitatively evaluate these approaches, the objective function is utilized as the error criterion for accuracy evaluation of the multi-view registration results. Table 1 compares the runtime and objective function value of the  registration results of the aforementioned approaches, where the bold number denotes the best performance among these competed approaches.
As shown in Table 1, the MA-ATICP algorithm can obtain almost the best performance in accuracy and efficiency among these approaches. The speed of MA-ATICP is much faster, especially in the case of big point cloud data.\par

\begin{table*}[t]
	\centering
	\caption{Performance comparisons based on different models.}
	\begin{tabular}{c c c c c c c}
		\hline
		
		\multirow{2}{*}{} 
        &  \multicolumn{2}{c}{LRS-Llalm} &  \multicolumn{2}{c}{MA-ICP}  &  \multicolumn{2}{c}{MA-ATICP}   \\ 
        \cline{2-3}        \cline{4-5}  \cline{6-7}
                           & Obj       & Time(s)   &Obj      &Time(s)  &Obj      &Time(s)   \\   
        \hline
         Bunny           &{0.6454}    & {50.2754} &{0.8533} &{301.998} &\bf {0.6434}  & \bf{40.942}\\
         Dragon         &0.4399      & 80.9986  &{0.5152}  &{251.916}     & \bf{0.4106}    & \bf{29.697}\\
         Happy           &0.137       & 321.4806  &{0.1821} &{1540.374}  &\bf{ 0.1363}    &\bf {128.634}\\      
	   \hline
		
	\end{tabular}
\label{tab:road_detection}
\end{table*}

\begin{figure}[!htbp]
  \includegraphics[width=11cm]{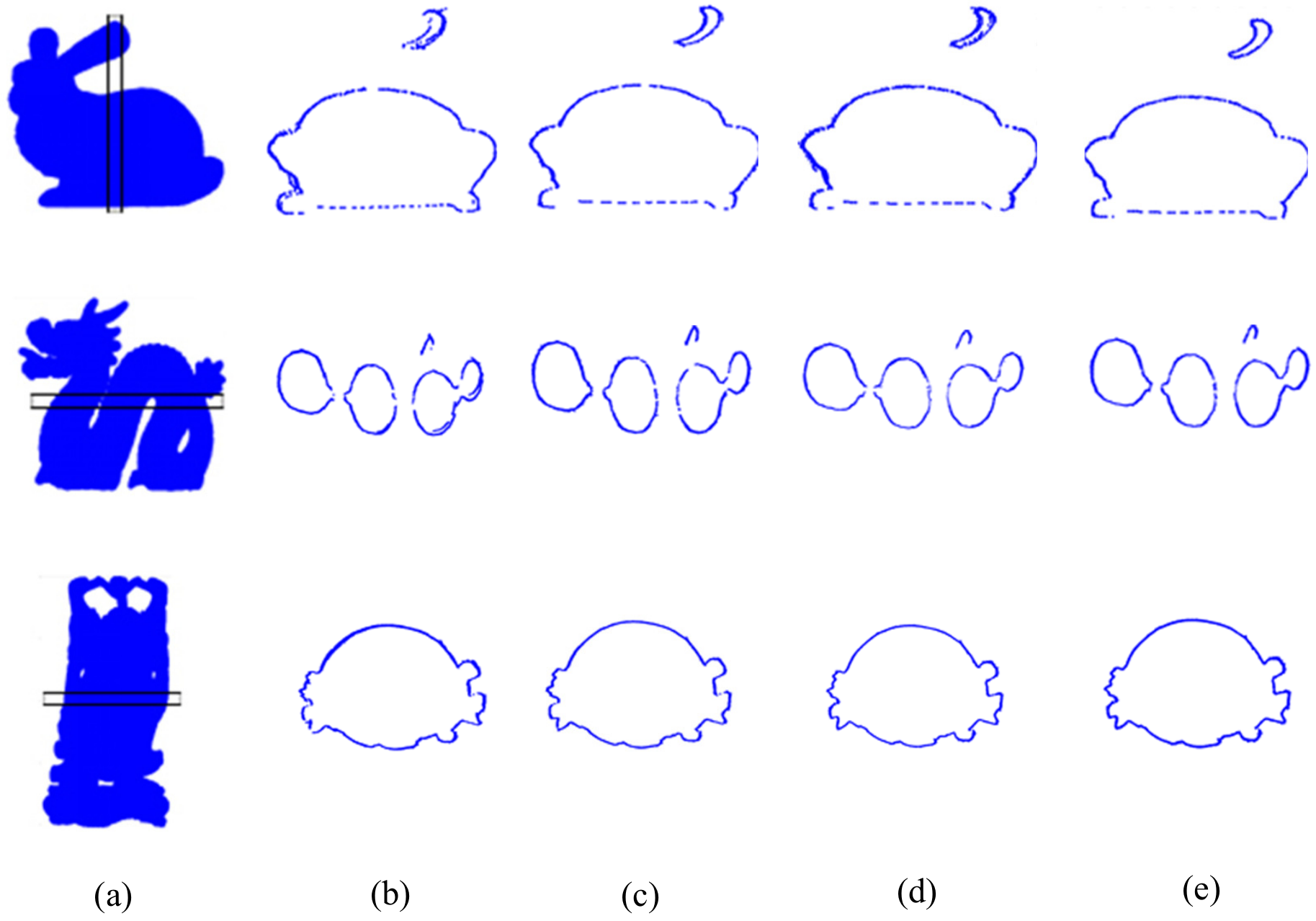}
\caption{Cross-section of multi-view registration results for two competed approaches. (a)3D model. (b)Cross-section of initial model. (c)Cross-section of LRS-Llalm. (d)Cross-section of MA-ICP. (e)Cross-section of MA-ATICP.}
\label{fig:Cross_section01}       
\end{figure}

\begin{figure}[!htbp]
  \includegraphics[width=11cm]{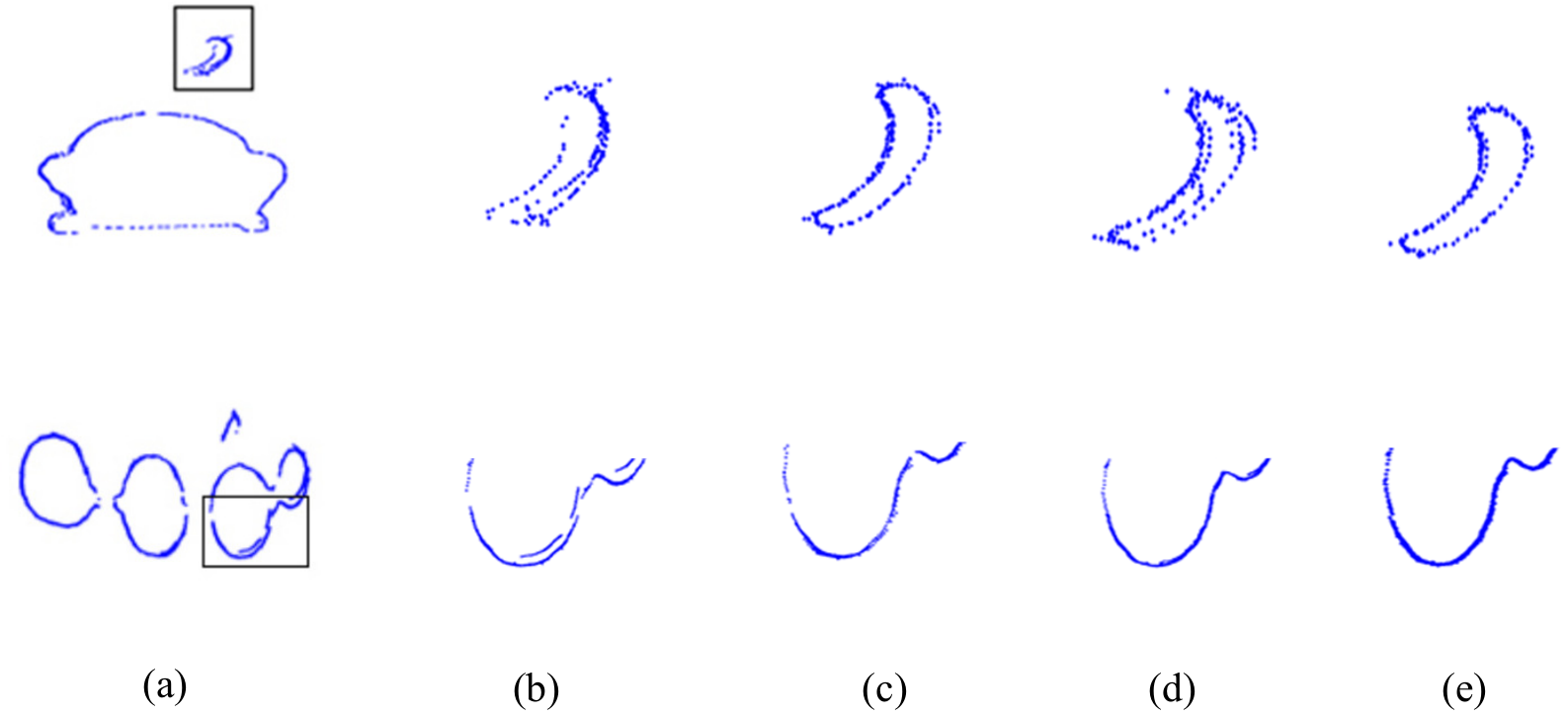}
\caption{Cross-section of partially amplified results for Bunny and Dragon. (a) Cross-section  of the initial model. (b)Amplified cross-section of the initial model. (c)Amplified cross-section of LRS-Llalm. (d)Amplified cross-section of MA-ICP. (e)Amplified cross-section of MA-ATICP. }
\label{fig:Cross_section02}       
\end{figure}

In order to evaluate the registration accuracy in a more intuitive way, Fig. \ref{fig:Cross_section01} shows the cross-sections of the three baseline methods based on the 3D models of Bunny, Dragon and Happy.
Fig. \ref{fig:Cross_section02} provides the cross-section of partially amplified results for comparison.
As shown in Fig. \ref{fig:Cross_section01} and Fig. \ref{fig:Cross_section02}, the MA-ATICP algorithm obtains the most accurate registration results over the other methods. \par

\begin{figure}[!htbp]
  \includegraphics[width=10cm]{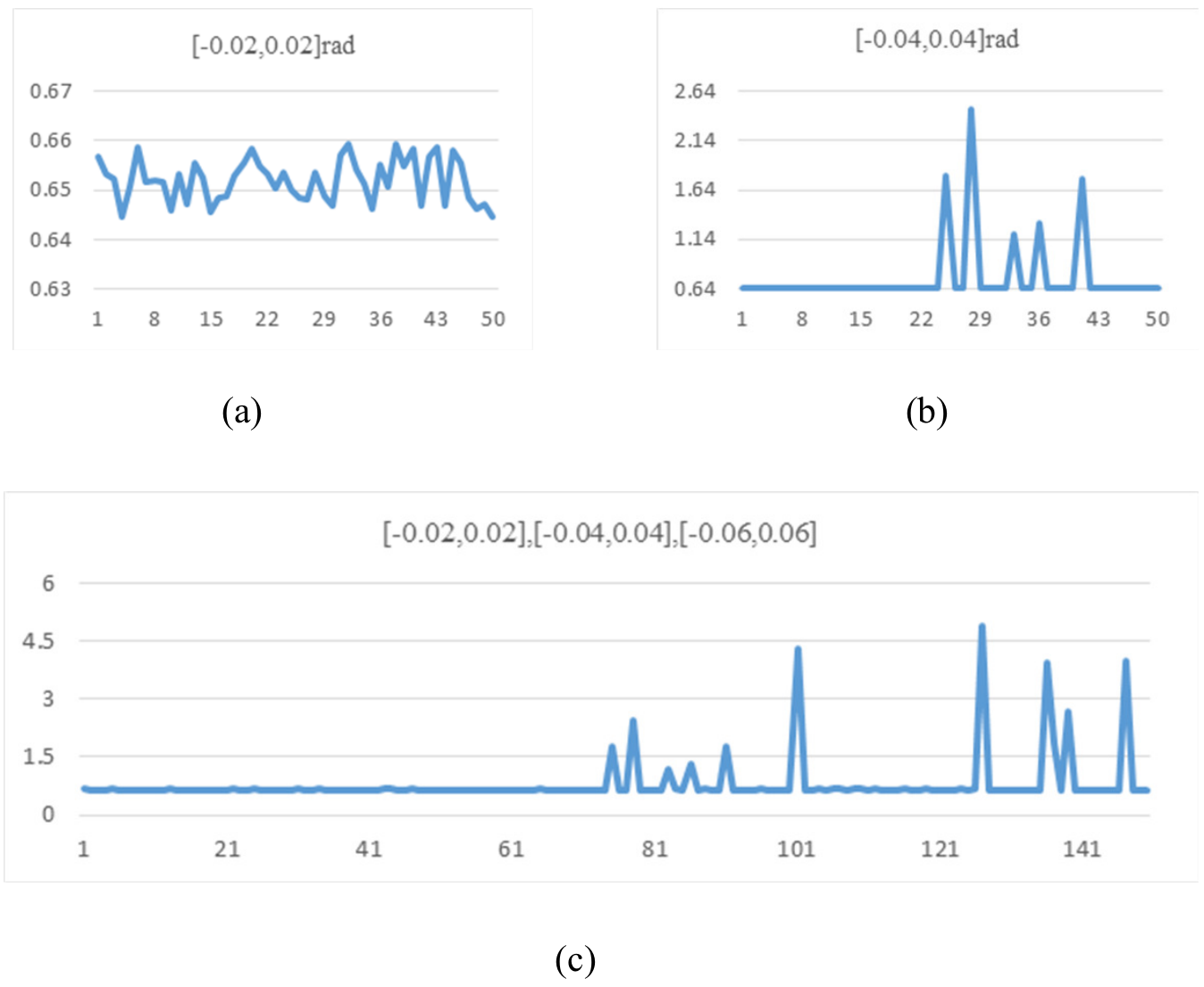}
\caption{The objective function value of the registration results for the LRS-Llalm in each MC trial, (a) and (b) are the amplified results of (c). }
\label{fig:Function_value_1}       
\end{figure}

\begin{figure}[!htbp]
  \includegraphics[width=10cm]{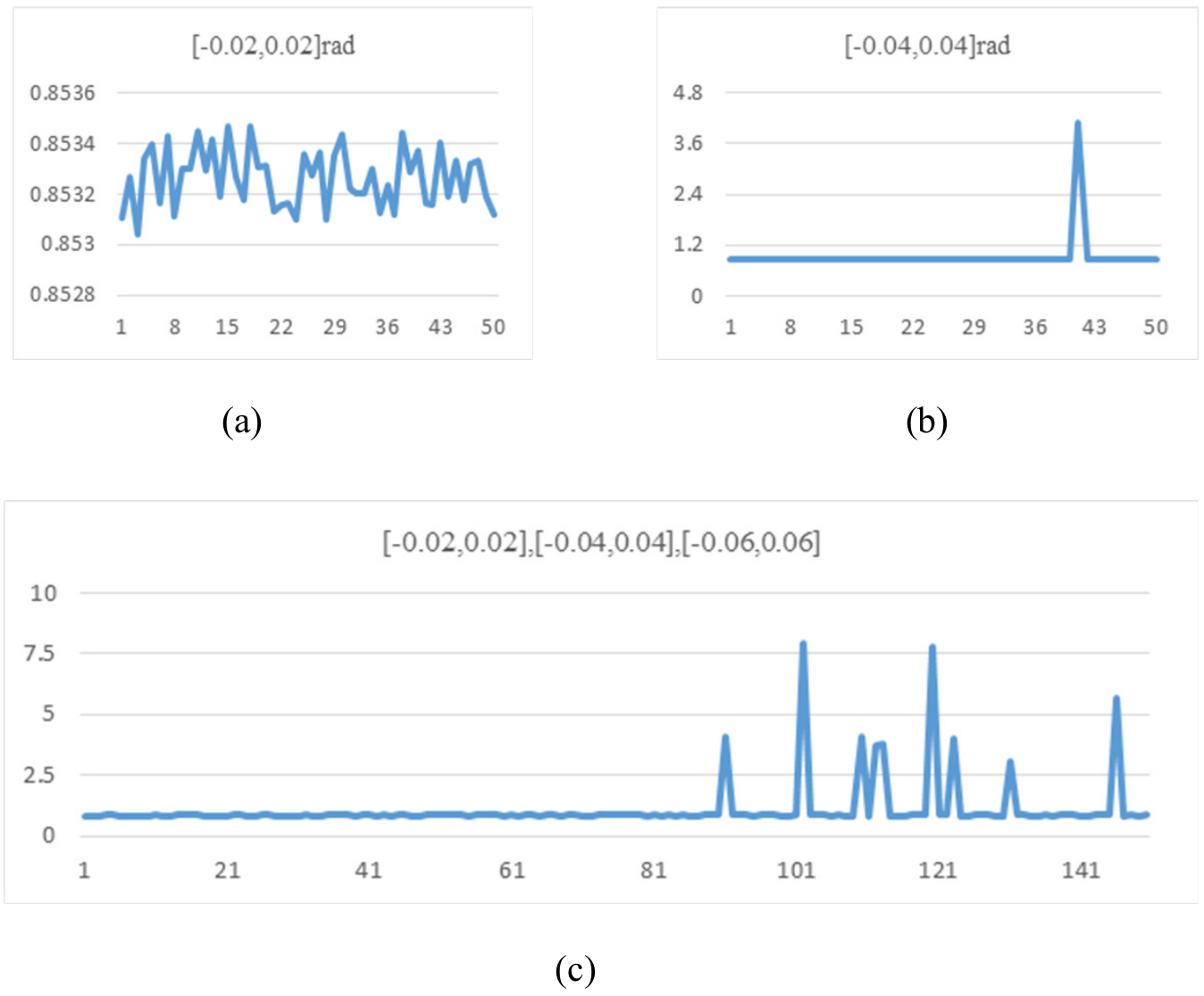}
\caption{The objective function value of the registration results for the MA-ICP in each MC trial, (a) and (b) are the amplified results of (c).}
\label{fig:Function_value_2}       
\end{figure}

\begin{figure}[!htbp]
  \includegraphics[width=10cm]{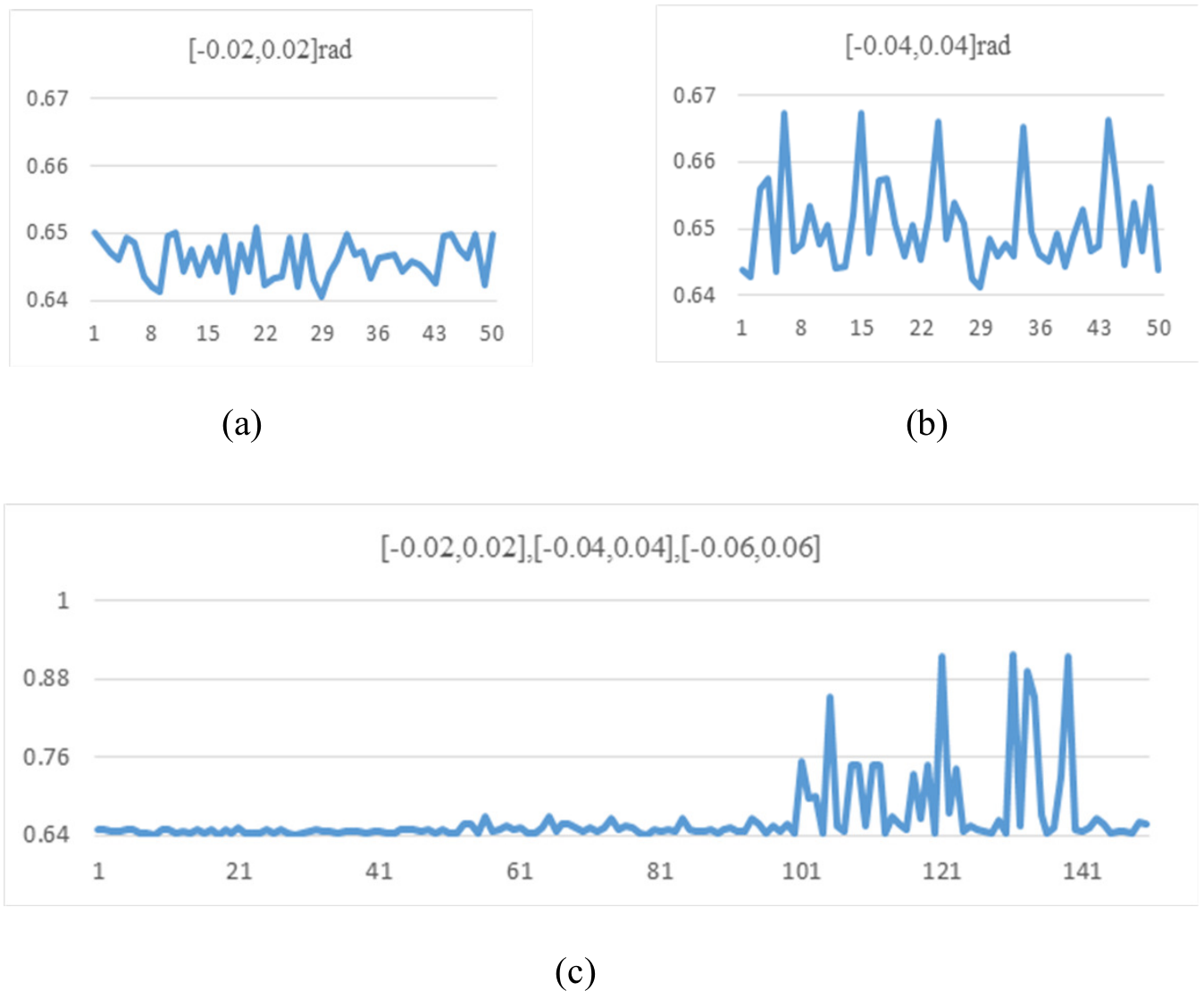}
\caption{The objective function value of the registration results for the MA-ATICP in each MC trial, (a) and (b) are the amplified results of (c).}
\label{fig:Function_value_3}       
\end{figure}

Three approaches are tested on Bunny with  initial parameters to evaluate the algorithm robustness.
The experiments are conducted by adding the uniform noises to the initial global motions $\{R_2^0, R_3^0, ..., R_N^0\}$. In order to eliminate randomness, we implement 50 Monte Carlo trials (MC trials)   with respect to three noise levels \cite{Guo_R}: $[-0.02,0.02]$, $[-0.04,0.04]$ and $[-0.06,0.06]$, as shown in Table 2.  
Mean(O) and Std(O) denote the mean accuracy and standard deviation of the objcts, while Mean(T) denotes the mean time cost.
The comparisons of mean value, standard deviation of objective function and the mean runtime are presented, where the bold numbers denote the best performance. 
To view the registration results in a more intuitive way, Fig. \ref{fig:Function_value_1}, Fig. \ref{fig:Function_value_2}  and Fig. \ref{fig:Function_value_3}  depict the the objective function value of the registration results for all the baseline methods in each MC trial. 
The comparison results demonstrate that the MA-ATICP algorithm gets the most accurate and robust registration results under different noise levels. \par
The LRS-Llalm algorithm incorporates the low-rank and sparse decomposition method to implement the multi-view registration, which can be failed due to the high ratio of missing relative motions. 
The MA-ICP algorithm adopts the traditional ICP algorithm to deal with registration between double-view  point cloud data with  partial overlap, and the ideal multi-view registration result cannot be obtained when the initial error is unendurable. 
In comparison with these baseline methods, the MA-ATICP algorithm reaches the best performance. \par

\begin{table}[]
	
	\caption{Performance comparisons under varied noise levels.}
\setlength{\tabcolsep}{0.8mm}{
	\begin{tabular}{c c c c c c c c c c}
		\hline
		
		\multirow{3}{*}{} 
        &  \multicolumn{3}{c}{$[-0.02,0.02]$rad}  &  \multicolumn{3}{c}{$[-0.04,0.04]$rad} &  \multicolumn{3}{c}{$[-0.06,0.06]$rad}  \\ 
        \cline{2-10}
       & Mean(O) &Std(O) &Mean(T) & Mean(O) &Std(O) &Mean(T) & Mean(O) &Std(O) &Mean(T)\\
        \hline
       LRS-Llalm &0.6500 &0.0048 &94.635 &0.6514  &0.3445 &104.857 &1.0227 &1.0071 &209.742 \\
       MA-ICP     &0.8532 &\bf{0.0002} &324.3  &0.9164  &0.4474 &312.582 &1.4980 &1.6478 &302.994 \\
       MA-ATICP  &\bf{0.6464} &0.0029 &\bf{20.149} &\bf{0.6505}  &\bf{0.0067} &\bf{21.696}  &\bf{0.6992} &\bf{0.0793} &\bf{25.477} \\
	   \hline
		
	\end{tabular}}
\label{tab:road_detection}
\end{table}

\begin{figure}[!htbp]
  \includegraphics[width=12cm]{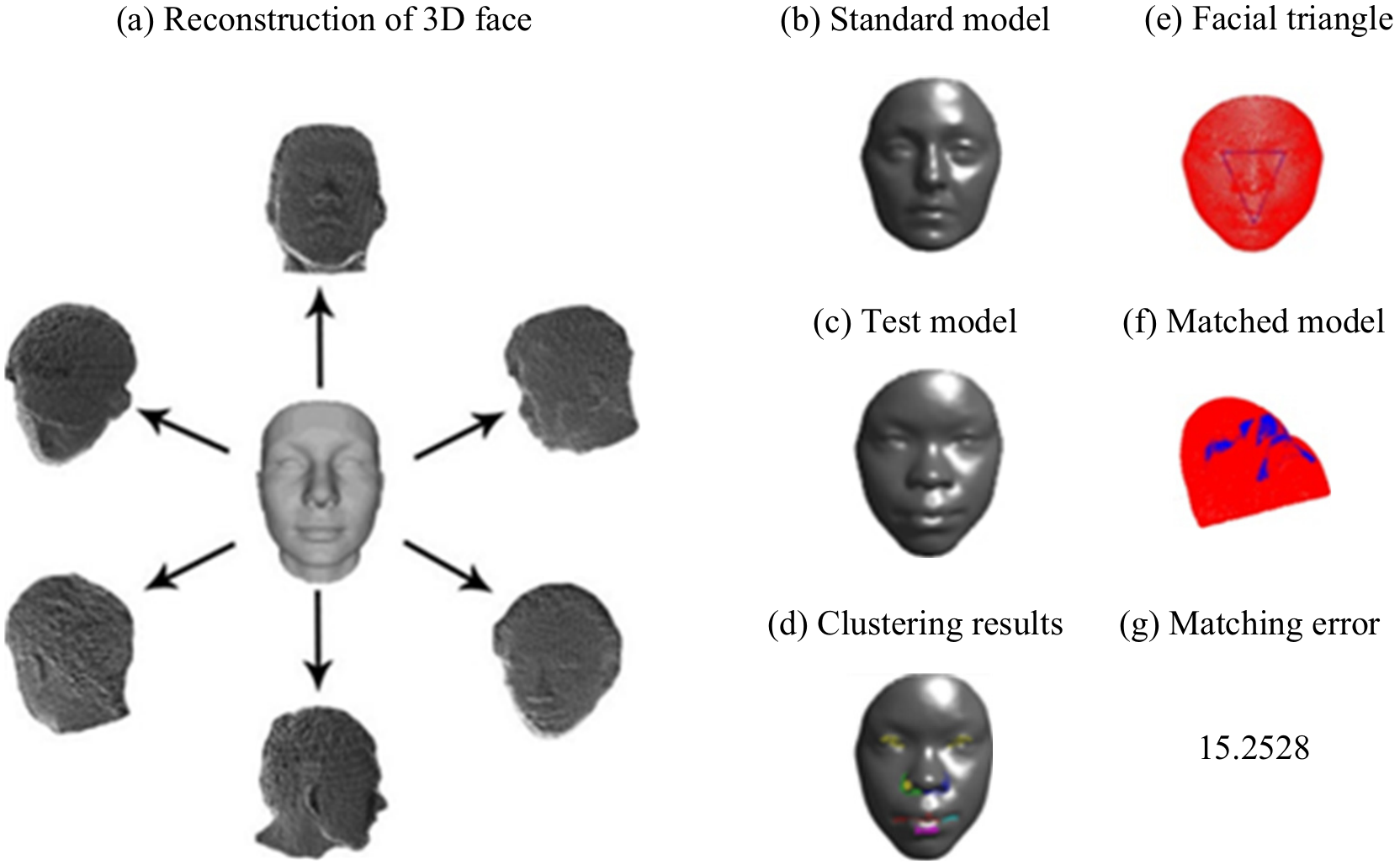}
\caption{Illustration of 3D face matching process.}
\label{fig:3D_face_detection}       
\end{figure}

\begin{table}[]
	
	\caption{The matching errors computed for 36 3D face models.}
	\begin{tabular}{c c c c c c }
		\hline
		index    &matching error  &index  &matching error  &index  &matching error\\
        \hline
1	&13.3297	&13	&6.6086	&25	&9.1297\\
2	&13.8684	&14	&9.3108	&26	&5.2139\\
3	&13.0567	&15	&8.8308	&27	&7.3424\\
4	&10.9594	&16	&10.6764	&28	&10.6764\\
5	&14.7189	&17	&10.8464	&29	&7.5995\\
6	&6.9949	    &18	&8.1302	&30	&18.2861\\
7	&10.8464	&19	&7.9781	&31	&11.5770\\
8	&13.9106	&20	&14.0030	&32	&8.8903\\
9	&8.6825	    &21	&5.8430	&33	&8.5329\\
10	&8.0042	    &22	&12.0190	&34	&6.1100\\
11	&7.8590	    &23	&14.5614	&35	&15.4480\\
12	&11.2252	&24	&11.9552	&36	&15.2528\\

	   \hline
		
	\end{tabular}
\label{tab:road_detection}
\end{table}

The registration of multi-view point cloud data lays a solid foundation for the application of 3D model retrieval. 
As a typical example of 3D model retrieval, we further evaluate the effectiveness of the 3D face detection. 
The reconstructed 3D face can be utilized as test data  for 3D face detection, and the matching error between the test and the standard models is computed, as shown in Fig. \ref{fig:3D_face_detection}.
Moreover, the matching errors for 36  test 3D faces are evaluated in Table 3. 
The face and non-face models can be easily discriminated by setting a proper threshold of matching error.\par

\section{Conclusion and Future Works}
In this paper, we propose a new algorithm for multi-view point cloud registration with adaptive convergence threshold. 
The point cloud registration is implemented based on an improved ICP algorithm and motion average algorithm. 
For the application of 3D model retrieval, we design a method for 3D face detection using geometric saliency. 
The test facial triangle is generated based on the saliency map, which is applied to compare with the standard facial triangle.
The proposed algorithm demands a certain overlap rate between the point cloud pairs.
The precise registration results will not be reached if the overlap rate is too small.
Besides, the registration of the multi-view point cloud data at different scales is a focus of further research.\par
In the future works, the array point cloud data will be studied instead of the laser point cloud.
The scale factors will be applied to specify the transformation relationship between the test and the benchmark point cloud data.
The overlap rate between the point cloud data  can be increased.
The partially overlapping point cloud data are then algined based on the optimal scaling ICP method.
For the application of 3D model retrieval, new vertex descriptors will be considered in addition to the curvature descriptor.
Moreover, various research based on 3D models will be studied, such as 3D building model, 3D vehicle model, etc.

\end{document}